\documentclass[11pt,a4paper]{article}

\usepackage[dvipsnames]{xcolor}
\usepackage[hyperref]{emnlp2020}

\usepackage{subcaption}
\usepackage{mwe}
\usepackage{booktabs} 
\usepackage{svrsymbols}

\usepackage{times}
\usepackage{latexsym}
\usepackage{graphicx} 
\usepackage{tikz-dependency}
\usepackage{tikz-qtree}

\aclfinalcopy 
\def\aclpaperid{2134} 

\title{LSTMS Compose---and Learn---Bottom-Up}

\author{Naomi Saphra \\
  University of Edinburgh \\
  \texttt{n.saphra@ed.ac.uk} \\\And
  Adam Lopez \\
  University of Edinburgh \\
  \texttt{alopez@inf.ed.ac.uk} \\}

\date{}

\begin{document}
\maketitle

\begin{abstract}
Recent work in NLP shows that LSTM language models capture hierarchical structure in language data. In contrast to existing work, we consider the \textit{learning} process that leads to their compositional behavior. For a closer look at how an LSTM's sequential representations are composed hierarchically, we present a related measure of Decompositional Interdependence (DI) between word meanings in an LSTM, based on their gate interactions. We connect this measure to syntax with experiments on English language data, where DI is higher on pairs of words with lower syntactic distance. To explore the inductive biases that cause these compositional representations to arise during training, we conduct simple experiments on synthetic data. These synthetic experiments support a specific hypothesis about how hierarchical structures are discovered over the course of training: that LSTM constituent representations are learned bottom-up, relying on effective representations of their shorter children, rather than learning the longer-range relations independently from children. 
\end{abstract}

\section{Introduction}


For years the LSTM dominated language architectures. It remains a popular architecture in NLP, and unlike Transformer-based models, it can be trained on small corpora~\citep{tran_importance_2018}.\footnote{As evidence of the ongoing popularity of LSTMs in NLP, a Google Scholar search restricted to \texttt{aclweb.org} since 2019 finds 191 citations to the original LSTM paper \citep{hochreiter1997long} and 242 citations to the original Transformer paper \citep{vaswani2017attention}.} \citet{abnar_transferring_2020} even found that the recurrent inductive biases behind the LSTM's success are so essential that distilling from them can improve the performance of fully attentional models. However, the reasons behind the LSTM's effectiveness in language domains remain poorly understood.

A Transformer can encode syntax using attention \citep{hewitt_structural_nodate}, and some LSTM variants explicitly encode syntax \citep{bowman-etal-2016-fast,DBLP:journals/corr/DyerKBS16}. So, the success of these models is partly explained by their ability to model syntactic relationships when predicting a word. By contrast, an LSTM simply scans a sentence from left to right, accumulating meaning into a hidden representation one word at a time, and using that representation to summarize the entire preceding sequence when predicting the next word. Yet we have extensive evidence that trained LSTMs are also sensitive to syntax. For example, they can recall more history in natural language data than in similarly Zipfian-distributed $n$-gram data, implying that they exploit linguistic structure in long-distance dependencies \citep{liu_lstms_2018}. Their internal representations appear to encode constituency \citep{blevins_deep_2018,hupkes_visualisation_2018} and syntactic agreement \citep{lakretz-etal-2019-emergence,gulordava-etal-2018-colorless}. In this paper, we consider how such representations are learned, and what kind of inductive bias supports them.

To understand how LSTMs exploit syntax, we use \textbf{contextual decomposition} (CD; Section~\ref{sec:cd}), a method that computes how much the hidden representation of an LSTM depends on particular past span of words. We then extend CD to \textbf{Decompositional Interdependence} (DI; Section~\ref{sec:interdependence}), a measure of interaction between spans of words to produce the representation at a particular timestep. For example, in the sentence ``Socrates asked the student trick questions’’, we might expect the hidden representation of the LSTM at the word ``questions’’ to interact primarily with its syntactic head ``asked’’, and less with the direct object ``the student''. If so, then an LSTM could be seen as implementing compositional \emph{localism} \citep{hupkes_compositionality_2020}: if a hidden representation encodes meaning, then this meaning is composed from local syntactic relationships. Our experiments on syntactically-parsed corpora (Section~\ref{sec:english}) illustrate this property --- interdependence decreases with syntactic distance, stratified by surface distance.

We then turn to a hypothesis about how such representations are learned. Using a simple synthetic corpus (Section~\ref{sec:long}), we allow LSTMs to learn to represent short sequences before they learn longer sequences that are dependent on them. Our goal is to then illustrate how they \textit{use} representations of short sequences in order to learn longer dependencies---if these smaller constituents are unfamiliar, LSTMs learn more slowly. Further experiments (Section~\ref{sec:isolating}) isolate hierarchical behavior from other factors causing local relations to be learned first, indicating that the model tends to build a subtree from its smaller constituents. We conclude that LSTMs \textit{compose} hierachically because they \textit{learn} bottom-up.
 
\section{Methods}\label{sec:methods}

Our DI measure is a natural extension of Contextual Decomposition~\citep[CD; ][]{murdoch_beyond_2018}, a tool for analyzing the representations produced by LSTMs. To conform with \citet{murdoch_beyond_2018}, our English language experiments use a one layer (400-dim) LSTM, with inputs taken from an embedding layer and outputs processed by a softmax layer.

\subsection{Contextual Decomposition} \label{sec:cd} 

We now will provide a blackbox explanation of CD, the groundwork for our DI. Let us say that we need to determine when our language model has learned that ``either'' implies  an appearance of ``or'' later in the sequence---a convenient test used since at least~\citet{1056813}. We consider an example sentence, ``\textit{Either} Socrates is mortal \textit{or} not''.  Because many nonlinear functions are applied in the intervening span ``Socrates is mortal'', it is difficult to directly measure the influence of ``either'' on the later occurrence of ``or''. To dissect the sequence and understand the impact of individual elements in the sequence, we could employ CD.

\begin{figure}
    \centering
    \includegraphics[width=0.48\textwidth]{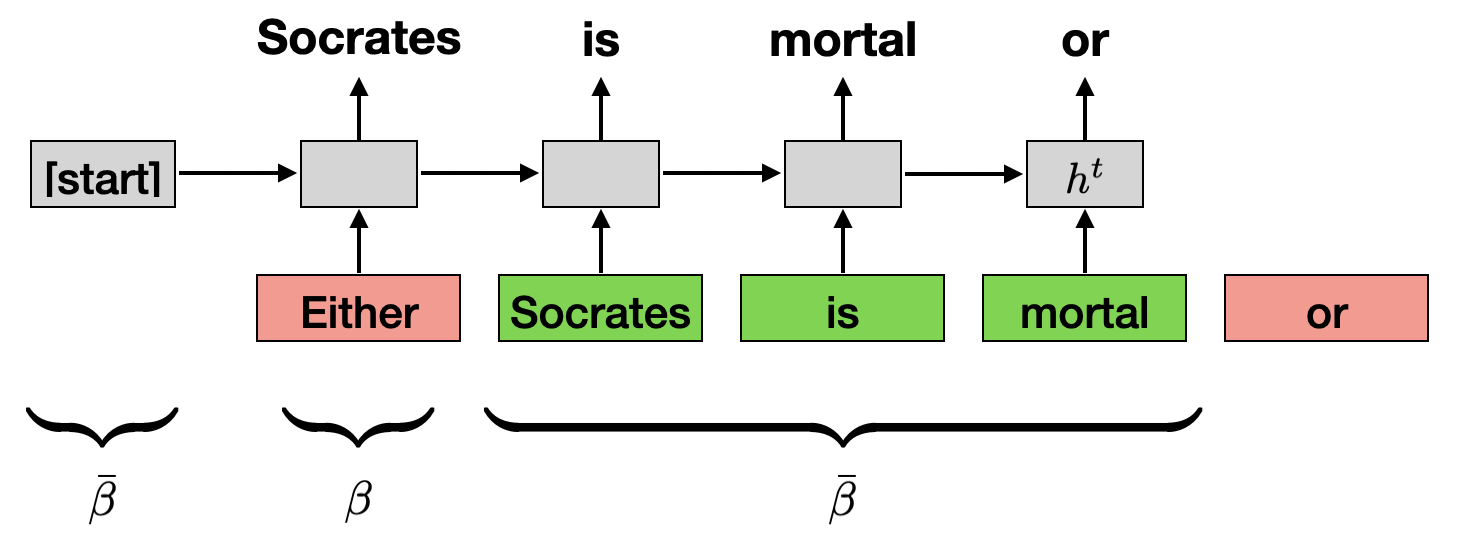}
    \caption{CD uses linear approximations of gate operations to linearize  the sequential application of the LSTM module. CD produces the vector $h^t_{\beta}$  isolating the contribution of ``Either'' to the vector $h^t$ predicting ``or'', as well as producing the irrelevant contribution $h^t_{\bar{\beta}; \beta \interaction \bar{\beta}}$. The irrelevant contribution considers both $\bar{\beta}$ and its interactions with $\beta$. In our figures, red will represent matched tokens and green the intervening span of tokens through which information must pass to predict the match.} 
    \label{fig:cd_example}
\end{figure}

CD is a method of looking at the individual influences that words and phrases in a sequence have on the output of a recurrent model. Illustrated in Figure~\ref{fig:cd_example},  CD decomposes the activation vector produced by an LSTM layer into a sum of relevant and irrelevant parts. The \textbf{relevant} part is the exclusive contribution of the set of words \textbf{in focus}, i.e., a set of words whose impact we want to measure. We denote this set of words as $\beta$. The \textbf{irrelevant} part includes the contribution of all words not in that set (denoted $\bar{\beta}$) as well as \textbf{interactions} between the relevant and irrelevant words (denoted $\beta \interaction \bar{\beta}$). For an output hidden state vector $h^t$, CD will decompose it into two vectors: the relevant $h^t_{\beta}$, and irrelevant $h^t_{\bar{\beta}; \beta \interaction \bar{\beta}}$, such that:
\begin{equation}
h \approx h^t_{\beta} + h^t_{\bar{\beta}; \beta \interaction \bar{\beta}}
\end{equation}
This decomposition of the hidden state is based on individual Shapley decompositions of the gating mechanisms themselves, as detailed in Appendix~\ref{sec:cd_details}.

Because the individual contributions of the items in a sequence interact in nonlinear ways, this decomposition is only an approximation and cannot exactly compute the impact of a  specific word or words on the  label predicted.  CD linearizes hidden states with low approximation error, but the presence of slight nonlinearities in the interactions between components forms the basis for our measure of Decompositional Interdependence later on.\footnote{In our analyses, CD yielded mean approximation error $\frac{\|(v^t_{\beta} + v^t_{\bar{\beta}; \beta \interaction \bar{\beta}})-v\|}{\|v\|}<10^{-5}$ at the logits. However, this measurement misses another source of approximation error: the allocation of credit between $\beta$ and the interactions  $\beta \interaction \bar{\beta}$. Changing the sequence out of focus $\bar{\beta}$ might influence $v^t_{\beta}$, for example, even though the contribution of the words in focus should be mostly confined to the irrelevant vector component. This approximation error is crucial because the component attributed to $\beta \interaction \bar{\beta}$ is central to our measure of DI.}

We can use softmax to convert the relevant logits (the hidden units after a linear transformation) $v^t_{\beta}$ into a probability distribution as $P(Y \mid x_{\beta}) = \textrm{softmax}(v^t_{\beta})$. This allows us to analyze the effect of input $x_{\beta}$ on the probability of a later element while controlling for the influence of the rest of the sequence. 

\subsection{Decompositional Interdependence} \label{sec:interdependence}

Next, we extend CD to focus on nonlinear interactions.  We frame compositionality in terms of whether the meanings of a pair of words or word subsets can be treated independently. For example, a ``slice of cake'' can be broken into the individual meanings of ``slice'', ``of'', and ``cake'', but an idiomatic expression such as ``piece of cake'', meaning a simple task, cannot be broken into the individual meanings of ``piece'', ``of'', and ``cake''. The words in the idiom likely have higher \textbf{Decompositional Interdependence}, or reliance on their interactions to build meaning. Another influence on DI should be syntactic relation; if you ``happily eat a slice of cake'', the meaning of ``cake'' does not depend on ``happily'', which modifies ``eat'' and is far on the syntactic tree from ``cake'', but the meaning of ``cake'' should be more dependent on ``slice'', which gives context for its part of speech and suggests that it is concrete.\footnote{In our natural language experiments, we focus on dependency relations, but the inductive bias we observe is towards broadly hierarchical patterns in which longer relations depend on local constituents. DI analysis of other sources of this latent hierarchical structure, such as idiom, are left to future work.} We will use the nonlinear interactions in contextual decomposition to analyze the DI between words alternately considered in focus.

Generally, CD considers all nonlinear interactions between the relevant and irrelevant sets of words to fall under $\beta \interaction \bar{\beta}$, the irrelevant contribution, although other allocations of interactions have been proposed~\cite{jumelet_analysing_2019}.
DI uses these nonlinearities to discover how strongly a pair of spans are associated.
A fully flat structure for building meaning could lead to a contextual representation that requires memorization of each word, breaking the simplifying assumption at the heart of CD that each word has an independent meaning to be incorporated into the sentence.


Given two interacting sets of words to potentially designate as the $\beta$ in focus, $A,B$ such that $A \cap B = \emptyset$, we use a measure of DI to quantify the degree to which $A \cup B$ be broken into their individual meanings. With $h^t_A$ and $h^t_B$ denoting the relevant contributions at the hidden layers of $A$ and $B$ according to CD, and $h^t_{A \cup B}$ as the relevant contribution of $A \cup B$, we compute the magnitude of nonlinear interactions, rescaled to control for the magnitude of the representation: 
\begin{equation}
\textit{DI}^t
(A,B) = \frac{\|h^t_{A \cup
B}-(h^t_{A}+h^t_{B})\|_2}{\|h^t_{A \cup B}\|_2}
\end{equation}

\begin{figure}\begin{center}
    \begin{dependency}[theme = simple]\small
        \begin{deptext}[column sep=1em]
          Socrates \& asked \& the \& student \& trick \& questions \\
        \end{deptext}
    \deproot[edge unit distance=2ex]{2}{ROOT}
    \depedge{2}{1}{\textsc{nsubj}}
    \depedge{2}{4}{\textsc{iobj}}
    \depedge{2}{6}{\textsc{obj}}
    \depedge{4}{3}{\textsc{det}}
    \depedge{6}{5}{\textsc{adj}}
    \end{dependency}\end{center}
    \caption{A dependency parsed sentence.} \label{fig:chimney}
\end{figure}

\begin{figure}
    \centering
    \includegraphics[width=0.48\textwidth]{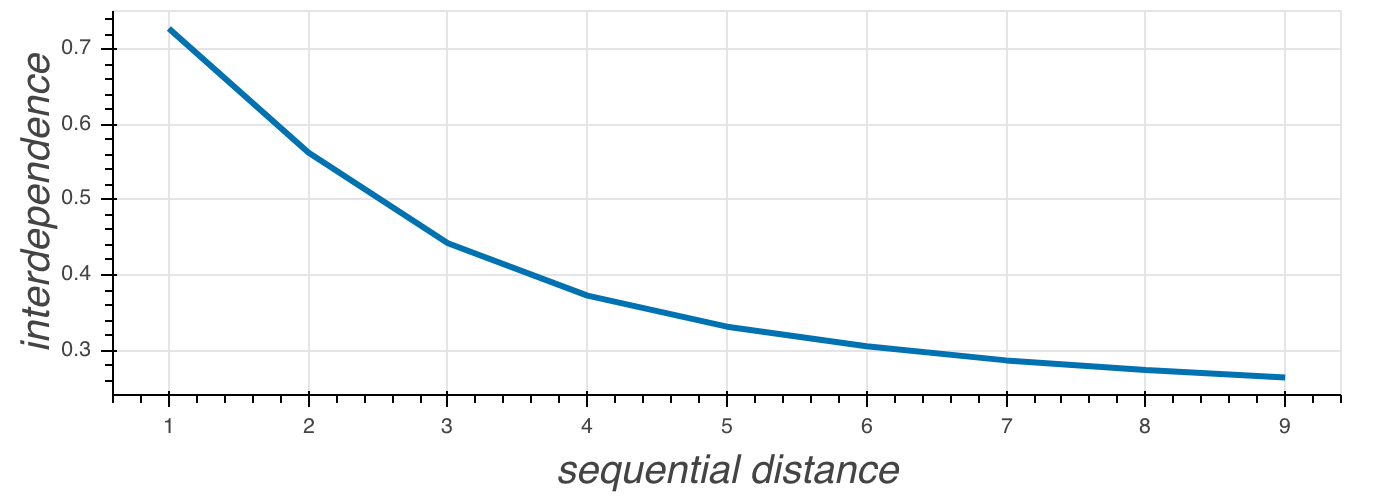}
    \caption{Average DI between word pairs $x_l, x_r$ at different sequential distances $r-l$.}
    \label{fig:interdep_all}
\end{figure}{}

\begin{figure*}
        \centering
        \begin{subfigure}[b]{0.18\textwidth}
            \centering
            \includegraphics[width=\textwidth]{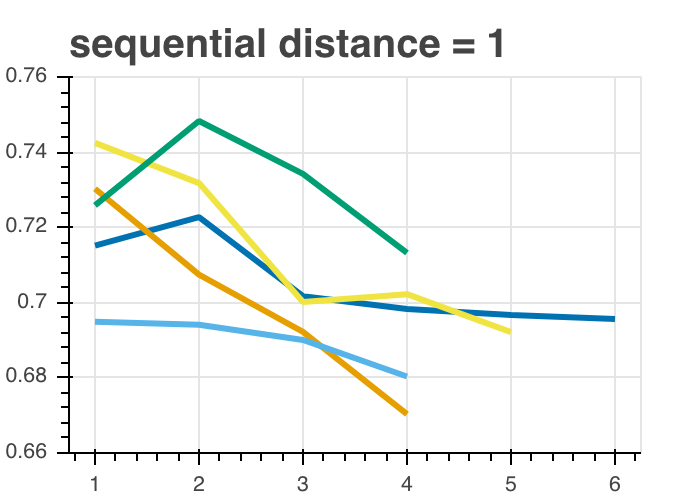}
            
        \end{subfigure}
        \hfill
        \begin{subfigure}[b]{0.18\textwidth}
            \centering
            \includegraphics[width=\textwidth]{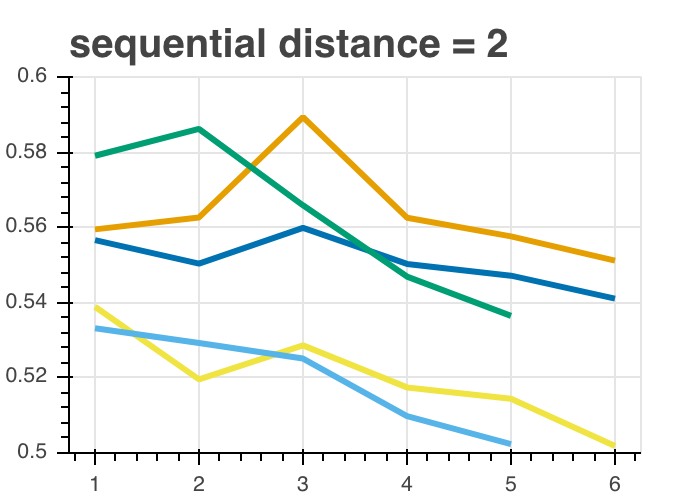}
            
        \end{subfigure}
        \hfill
        \begin{subfigure}[b]{0.18\textwidth}
            \centering
            \includegraphics[width=\textwidth]{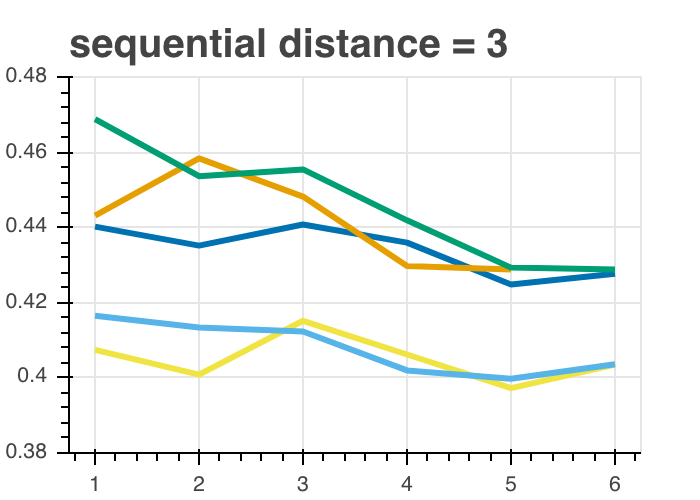}
            
        \end{subfigure}
        \hfill
        \begin{subfigure}[b]{0.18\textwidth}
            \centering
            \includegraphics[width=\textwidth]{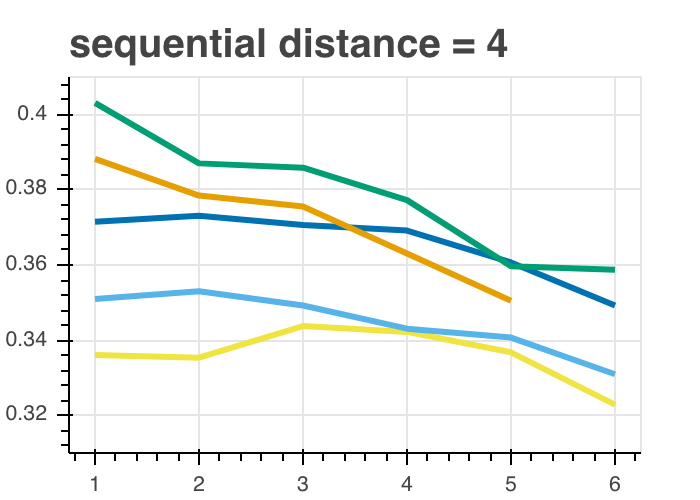}
            
        \end{subfigure}
        \hfill
        \begin{subfigure}[b]{0.18\textwidth}  
            \centering 
            \includegraphics[width=\textwidth]{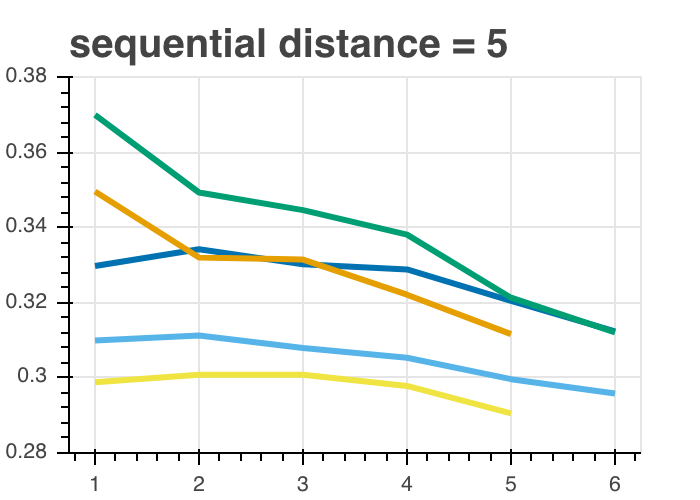}
            
        \end{subfigure}
        \vskip\baselineskip
        \begin{subfigure}[b]{0.18\textwidth}
            \centering
            \includegraphics[width=\textwidth]{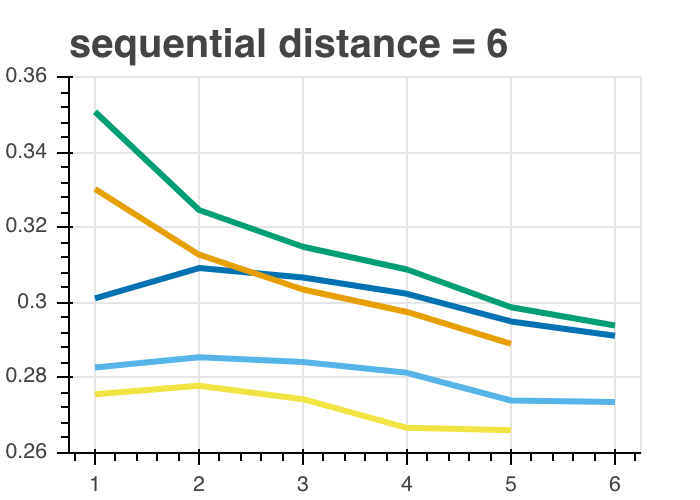}
        \end{subfigure}
        \hfill
        \begin{subfigure}[b]{0.18\textwidth}
            \centering
            \includegraphics[width=\textwidth]{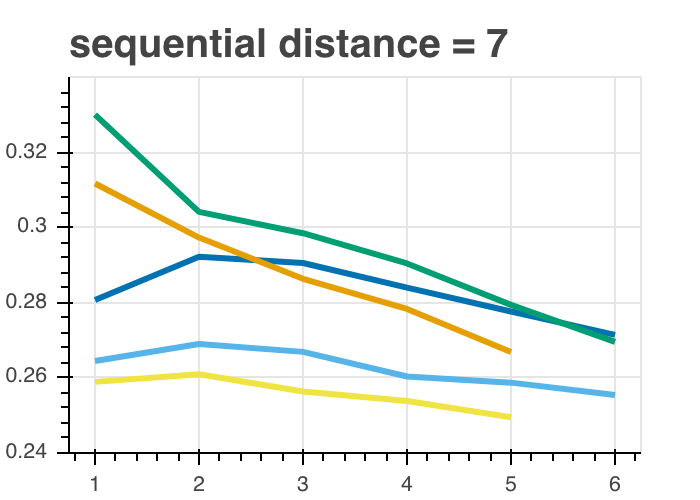}
        \end{subfigure}
        \hfill
        \begin{subfigure}[b]{0.18\textwidth}
            \centering
            \includegraphics[width=\textwidth]{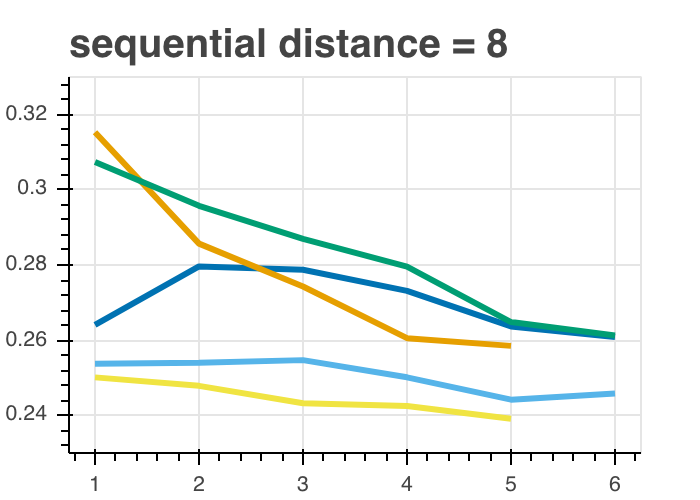}
        \end{subfigure}
        \hfill
        \begin{subfigure}[b]{0.18\textwidth}
            \centering
            \includegraphics[width=\textwidth]{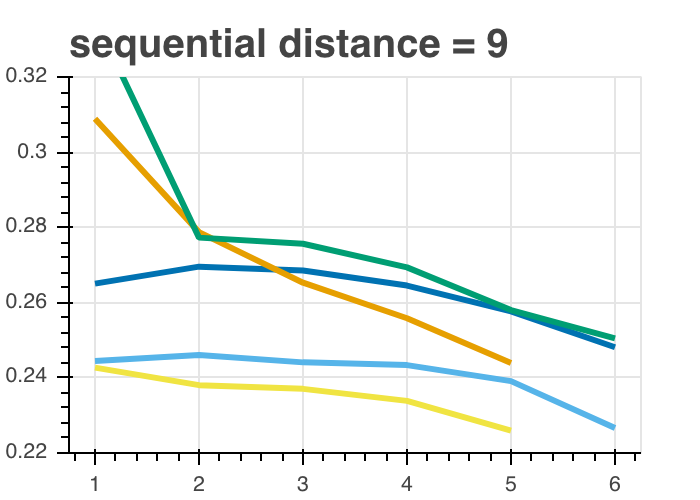}
        \end{subfigure}
        \hfill
        \begin{subfigure}[b]{0.18\textwidth}  
            \centering 
            \includegraphics[width=\textwidth]{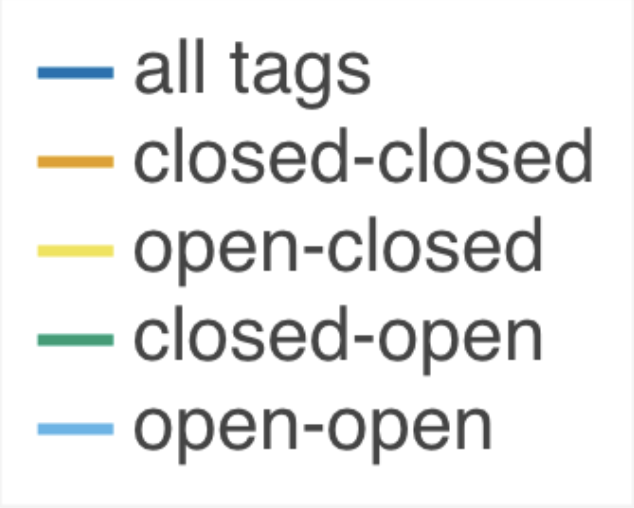}
        \end{subfigure}
\caption{Mean \textit{DI} (y-axis) between word pairs at varying \textit{syntactic distances} (x-axis), stratified by \textit{whether the POS tags are closed or open class} (line color) and by \textit{sequential distance} (plot title). The y-axis ranges differ, but the scale is the same for all plots. Each mean is plotted only if there are at least 100 cases to average.}
\label{fig:closed_and_open_tags}
\end{figure*}

This quantity is related to probabilistic independence. We would say that random variables $X$ and $Y$ are independent if their joint probability $P(X,Y) = P(X)P(Y)$. Likewise, the meanings of $A$ and $B$ can be called independent if $h^t_{A \cup B}=h^t_{A}+h^t_{B}$. A parallel can also be drawn to Information Quality Ratio~\citep{jetka_information-theoretic_2019}, a normalized form of mutual information which quantifies information exchanged between two variables against total uncertainty, if we view a decomposed output vector $h^t_{\beta}$ as information transmitted from $\beta$:
\begin{equation}
\textit{IQR}^t(A,B) = \frac{H(A,B) - H(A|B) - H(B|A)}{H(A,B)}
\end{equation}
    
Note that CD is applied to the representation at a particular timestep, and therefore DI is implicitly an operation that takes three parameters (excluding the sentence): $A, B$ and the timestamp at which to access their representations. However, in order to minimize information degradation over time, we access $h^t$ at the lowest timestep accommodating all spans in focus, $t = \textrm{max}(\textrm{idx}(A), \textrm{idx}(B))$.

Concurrently with this work, \citet{chen_generating_2020} also developed a method of studying the interaction between words using Shapley-based techniques like CD. However, their method was based on an assumption of underlying hierarchical structure and therefore unsuitable for the experiments we are about to conduct. Their results nonetheless validate the relationship between feature interaction and syntactic structure.

\section{English Language Experiments} \label{sec:english}

We now apply our measure of DI to a natural language setting to see how LSTMs employ bottom-up construction. In natural language, disentangling the meaning of individual words requires contextual information which is hierarchically composed. For example, in the sentence, ``Socrates asked the student trick questions'', ``trick questions'' has a clear definition and strong connotations that are less evident in each word individually. However, knowing that ``trick'' and ``student'' co-occur is not sufficient to clarify the meaning and connotations of either word or compose a shared meaning.

Here, we consider whether the LSTM observes headedness, by composing meaning between a headword and its immediate modifiers---behavior which a Recurrent Neural Network Grammar~\citep[RNNG;][]{DBLP:journals/corr/DyerKBS16} also learns~\citep{smith_what_2017}. If a standard LSTM learns similar behavior in line with syntax, it is \textit{implicitly} a syntactic language model.

These experiments use language models trained on wikitext-2~\cite{merity2016pointer}, run on the Universal Dependencies corpus English-EWT~\cite{silveira14gold}.

\subsection{DI and Syntax}

To assess the connection between DI and syntax, we consider the DI of word pairs with different syntactic distances. For example, in Figure~\ref{fig:chimney}, ``trick'' is one edge away from ``questions'', two from ``asked'', and four from ``the''. In Figure~\ref{fig:interdep_all}, we see that in general, the closer two words occur in sequence, the more they influence each other, leading to correspondingly high DI. Therefore we stratify by the sequential distance of words when we investigate syntactic distance.

As synthetic data experiments will show (Section~\ref{sec:synthetic}), phrase frequency and predictability play a critical role in determining DI (although we found raw word frequency shows no clear correlation with DI in English). In Figure~\ref{fig:closed_and_open_tags}, we control for these properties through stratifying by open and closed POS tag class. Open class POS tags frequently accept new words (e.g., nouns and adjectives), whereas closed class tags are mostly consistent historically (e.g., determiners and prepositions). These classes vary in their predictability in context; for example, determiners are almost always soon followed by a noun, but adjectives appear in many constructions like ``Socrates is mortal'' where they are not. Irrespective of both sequential distance and POS class, we see broadly decreasing trends in DI as the syntactic distance between words increases, consistent with the prediction that syntactic proximity drives DI. This pattern is clearer as words become further apart in the sequence, likely due to the absence of localized non-syntactic influences such as priming effects.

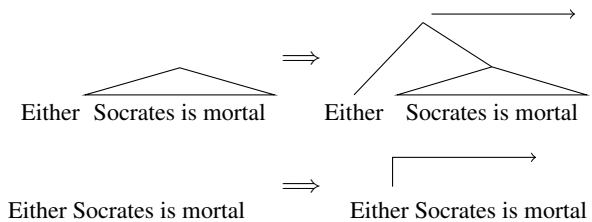
\begin{figure}\footnotesize
\begin{tikzpicture}
\begin{scope}[level distance=8mm]
\Tree 
[.\node(root){~}; \edge[roof]; \node(sim){Socrates is mortal}; ] 
\node[left of=sim,node distance=17mm]{Either};
\end{scope}

\node[right of=root, node distance=16mm](lra){$\Longrightarrow$};

\node[below of=lra, node distance=17mm]{$\Longrightarrow$};

\begin{scope}[xshift=32mm,yshift=6mm,frontier/.style={distance from root=14mm},level distance=7mm]
\Tree [.\node(root){~};
 Either 
[ \edge[roof]; {Socrates is mortal} ] ] 
\draw[->] (root) edge (20mm,0);
\end{scope}

\begin{scope}[xshift=38mm,yshift=-20mm]
\node(esim){Either Socrates is mortal};

\node[left of=esim, node distance=45mm]{Either Socrates is mortal};
\node[above of=esim,yshift=-8mm,xshift=-10mm](astart){};
\node[above of=esim,,yshift=-3mm,xshift=10mm](aend){};

\draw[->] (astart) |- (aend);
\end{scope}
\end{tikzpicture}

\caption{Top: A familiar span (indicated by a triangle illustrating it as a recognizable constituent) is used as a scaffold in its new context, allowing the model to construct a closely interdependent representation for predicting the next word. Bottom: An unfamiliar span cannot be used as a scaffold, so the model is forced to learn the either/or relation independently.}\label{fig:scaffolding}
\end{figure}

This behavior shows a tendency towards \textbf{hierarchical construction} aligned with syntax, wherein the LSTM ties a head's representation together with its child constituents and further associations are less dependent on each other. Similar behavior is the goal of RNNGs and other models which use stack LSTMs~\citep{dyer-etal-2015-transition}, which ensure the words in a constituent will be highly interdependent in their shared representation because the constituent will be based on a dictionary lookup for its subtree structure. In an RNNG, this behavior is a result of \textbf{bottom-up learning} during training, when the composition operation combines existing tag subtrees into a new lookup key. Our next experiments will illustrate how LSTMs already learn bottom-up implicitly, because they are biased towards the top behavior in Figure~\ref{fig:scaffolding} when a scaffolding environment is available.


\section{Synthetic Experiments} \label{sec:synthetic}

    \begin{figure*}
        \centering
        \begin{subfigure}[b]{0.48\textwidth}
            \centering
            \includegraphics[width=\textwidth]{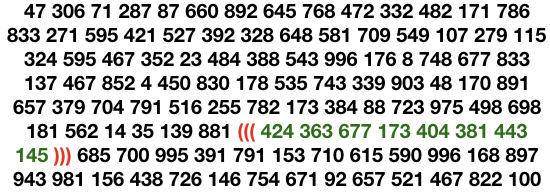}
            \caption[Network2]%
            {{\small unfamiliar-scaffold training set}}    
            \label{fig:unpred}
        \end{subfigure}
        \hfill
        \begin{subfigure}[b]{0.48\textwidth}  
            \centering 
            \includegraphics[width=\textwidth]{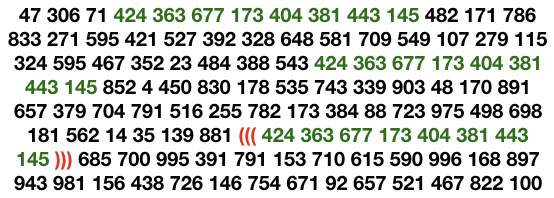}
            \caption[]%
            {{\small familiar-scaffold training set}}    
            \label{fig:pred}
        \end{subfigure}
        \vskip\baselineskip
        \begin{subfigure}[b]{0.48\textwidth}   
            \centering 
            \includegraphics[width=\textwidth]{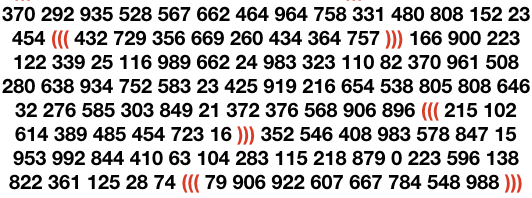}
            \caption[]%
            {{\small out-domain test set}}    
            \label{fig:outdo}
        \end{subfigure}
        \hfill
        \begin{subfigure}[b]{0.48\textwidth}   
            \centering 
            \includegraphics[width=\textwidth]{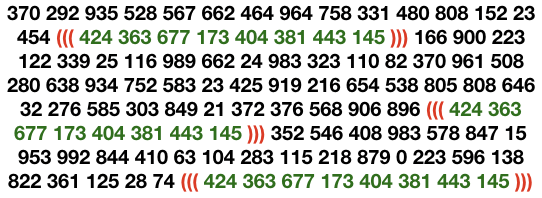}
            \caption[]%
            {{\small in-domain test set}}    
            \label{fig:indo}
        \end{subfigure}
        \caption{Caricatured train and test datasets for exploring the effect of scaffold familiarity on learning longer distance relations. We have highlighted rule boundaries {\color{red}$\alpha$} and {\color{red}$\omega$} in red, and scaffold {\color{ForestGreen}$q$} $ \in Q_k$ in green.}
    \end{figure*}
    
Our next experiments use synthetic data to show how training is bottom-up. LSTM training sees long-range connections discovered after short-range connections; in particular, document-level content topic information is encoded much later in training than local information like part of speech~\citep{saphra-lopez-2019-understanding}.

These experiments explain such learning phases by showing that the training process is inherently compositional due to bottom-up learning.\footnote{Other phenomena contribute but are outside our current focus. First, long-range connections are less consistent (particularly in a right-branching language like English), and will thus take longer to learn~(Appendix~\ref{sec:frequency}. For example, the pattern of a determiner followed by a noun will appear very frequently, as in ``the man'', while long-range connections like ``either/or'' are rarer. Second, rarer patterns are learned slowly  due to vanishing gradients~(Appendix~\ref{sec:bptt}).} 
That is, not only are the shorter sequences learned first, but they \textit{form the basis} for longer relations learned over them. For example, the model might learn to represent sequences like ``Socrates is mortal'' before it can learn to represent the either/or relation around it, building from short constituents to long. This behavior is seen in shift-reduce parsers and their neural derivatives like RNNGs.

Bottom-up training is not a given and must be verified.\footnote{In fact, learning simple rules early on might inhibit the learning of more complex rules through gradient starvation \cite{combes_learning_2018}, in which more frequent features dominate the gradient directed at rarer features.  Shorter familiar patterns could slow down the process for learning longer range patterns by trapping the model in a local minimum which makes the long-distance rule harder to reach.} However, if the hypothesis holds and training builds syntactic patterns hierarchically, it can lead to representations that are built hierarchically at inference time, reflecting linguistic structure, as we have seen. To test the idea of a compositional training process, we use synthetic data that controls for the consistency and frequency of longer-range relations. We find:

\begin{enumerate}
\item LSTMs trained with familiar intervening spans have poor performance predicting long distance dependents like ``or'' without familiar intervening spans (Figure~\ref{fig:speed}). This could be explained by the idea that they never acquire the either/or rule (instead memorizing the entire sequence).
\item But in fact, the either/or rule is acquired \textit{faster} with familiar constituents, as is clear even if the role of ``either'' is isolated (Figure~\ref{fig:cd_alpha}).
\item The poor performance is instead connected to high interdependence between ``either'' and the intervening span (Figures~\ref{fig:incremental} and \ref{fig:interdependence}).
\item Observations (2) and (3) support the idea that acquisition is biased towards bottom-up learning, using the constituent as a scaffold to support the long-distance rule.
\end{enumerate}




\subsection{Training Procedure}

We train our one-layer 200-dim LSTM with a learning rate set at 1 throughout and gradients clipped at 0.25. We found momentum and weight decay to slow rule learning in this setting, so they are not used.

\subsection{Long Range Dependencies} \label{sec:long}


First, we describe long-range rules whose acquisition will illuminate compositional learning dynamics. Consider how ``either'' predicts ``or'', often interceded by a closed constituent. To learn this rule, a language model must backpropagate information from the occurrence of ``or'' through the intervening span of words, which we will call a \textbf{scaffold}. Perhaps the scaffold is recognizable as a particular type of constituent: in ``Either  \textit{Socrates is  mortal} or not'', ``or'' becomes predictable after a constituent closes. But what if the scaffold is unfamiliar and its structure cannot be effectively represented by the model? For example, if the scaffold includes unknown tokens: ``Either \textit{slithy toves gyre} or not''. How will the gradient carried from ``or'' to ``either'' be shaped according to the scaffold, and how will the representation of that long-range connection change accordingly? 

A \textbf{familiar} scaffold like ``Socrates is mortal'' could be used by a bottom-up training process as a short constituent on which to build longer-range representations, so the meaning of ``Either'' will depend on a similar constituent. Conversely, if training is not biased to be compositional, the connection will be made regardless of the scaffold\footnote{Such behavior does reflect another aspect of compositionality, that of \textit{systematicity}~\citep{hupkes_compositionality_2020}.}, so the rule will generalize to test data: ``either'' will always predict ``or''. This either/or association might \textit{later} develop a dependency on the intervening span due to the nature of the data, but it will initially learn to predict without such scaffolding. We use a synthetic corpus to test these predictions.

In our synthetic corpus, we generate data uniformly at random from  a vocabulary $\Sigma$. We insert $n$ instances of the long-distance rule $\alpha \Sigma^k \omega$, with scaffold $\Sigma^k$ of length $k$, \textbf{open symbol} $\alpha$, and \textbf{close symbol} $\omega$, with $\alpha, \omega \not\in \Sigma$ (with $\alpha$ as ``either'' and $\omega$ as ``or''). Relating to our running example, $\alpha$ stands for ``either'' and $\omega$ stands for ``or''. We use a corpus of 1m tokens with $|\Sigma|=$ 1k types, which  leaves a low probability that any scaffold sequence longer than 1 token appears elsewhere by chance.

\begin{figure*}
        \centering
        \begin{subfigure}[b]{0.48\textwidth}
            \centering
            \includegraphics[width=\textwidth]{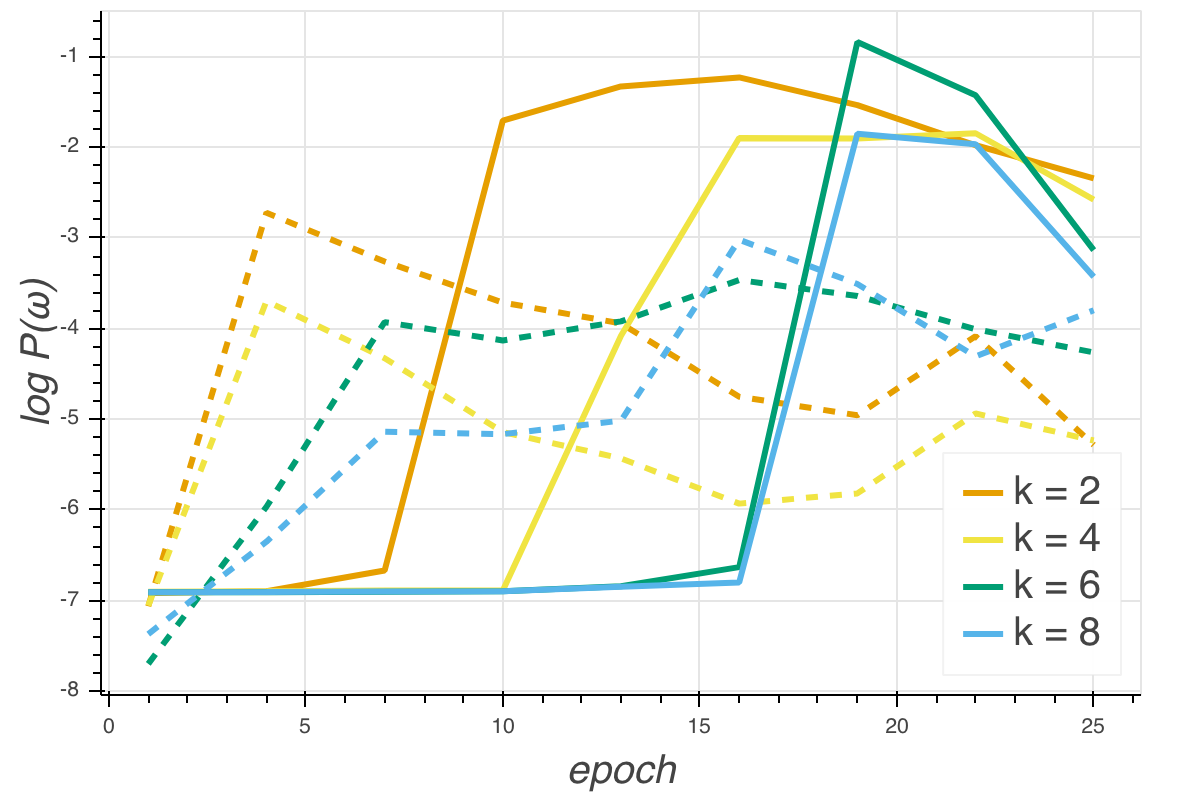}
            \caption[Network2]%
            {{\small In-domain scaffold test setting}}    
            \label{fig:indomain}
        \end{subfigure}
        \hfill
        \begin{subfigure}[b]{0.48\textwidth}  
            \centering 
            \includegraphics[width=\textwidth]{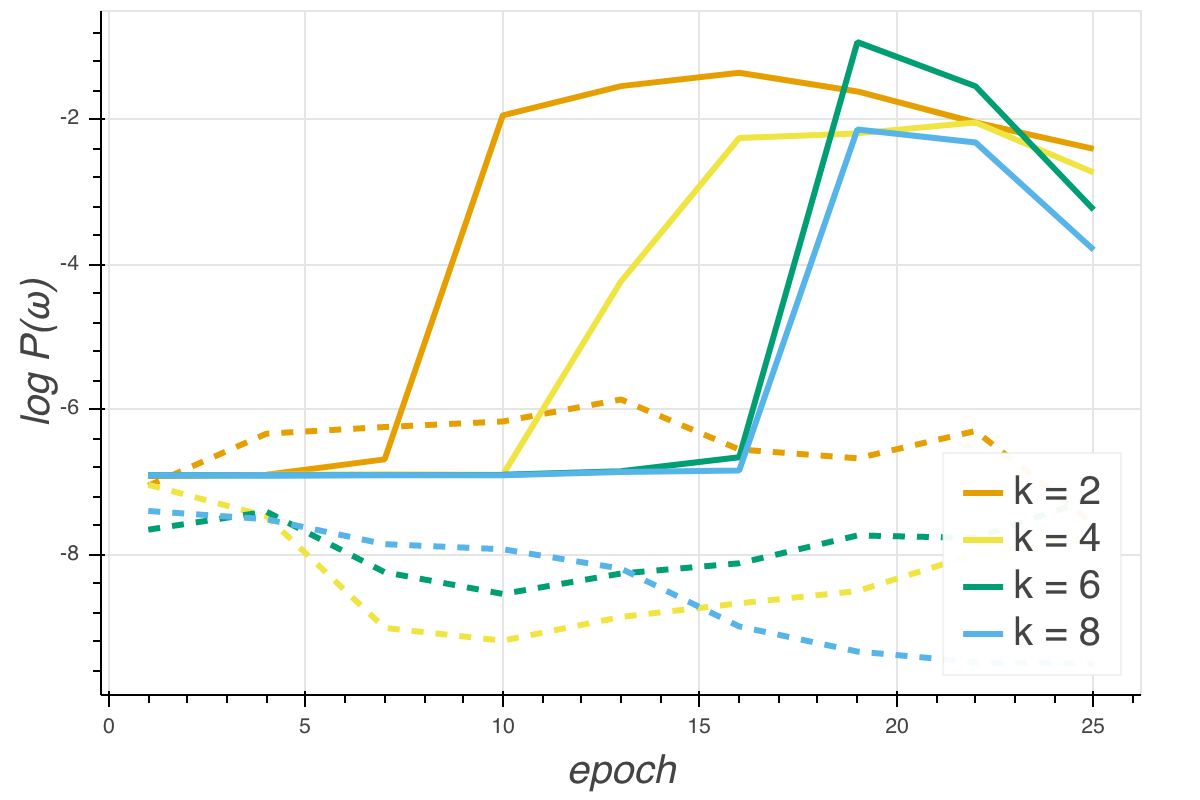}
            \caption[]%
            {{\small Random scaffold test setting}}    
            \label{fig:outdomain}
        \end{subfigure}
\caption{Mean marginal target probability of the close symbol in a rule. Solid lines are trained in the unfamiliar-scaffold set, dashed lines on familiar-scaffold. Color is specified by scaffold length ($k$). Scale of y-axis is matched among graphs.}
\label{fig:speed}
\end{figure*}

\subsection{The Effect of Scaffold Familiarity} \label{sec:familiarity}

To create a dataset of long-range connections with predictable scaffolds, we modify the original synthetic data (Figure~\ref{fig:unpred}) so each scaffold appears frequently outside of the $\alpha/\omega$ rule (Figure~\ref{fig:pred}). The scaffolds are sampled from a randomly generated vocabulary of 100 phrases of length $k$, so each unique scaffold $q$ appears in the training set 10 times in the context $\alpha q\omega$. This repetition is necessary in order to fit $1000$ occurrences of the rule in all settings. 

In the \textbf{familiar-scaffold setting}, we randomly distribute $1000$ occurrences of each scaffold throughout the corpus outside of the rule patterns. Therefore each scaffold is seen often enough to be memorized (see Appendix~\ref{sec:frequency}). In the original \textbf{unfamiliar-scaffold setting}, $q$ appears only as a scaffold, so it is not memorized independently. 

We also use two distinct test sets. Our in-domain test set (Figure~\ref{fig:indo}) uses the same set of scaffolds as the train set. In Figure~\ref{fig:indomain}, the model learns to predict the close symbol faster if the scaffolds are otherwise memorized. However, this effect may be due to vanishing gradients, discussed below.

These familiar scaffolds do not teach the general long distance dependency rule. If the test set scaffolds are sampled uniformly at random (Figure~\ref{fig:outdo}), Figure~\ref{fig:outdomain} shows that the familiar-scaffold training setting never teaches the model to generalize the $\alpha/\omega$ rule. For a model trained on the familiar domain, a familiar scaffold is required to predict the close symbol.

\paragraph{Vanishing Gradients:} A familiar intervening span is predictably a less effective scaffold, because the familiarity will limit longer distance information due to vanishing gradients. Consider in a simple RNN, as the gradient of the error $e^t$ at timestep $t$ backpropagates $k$ timesteps through the hidden state $h$:
$$\frac{\partial e^t}{\partial h_{t-k}} = \frac{\partial e^t}{\partial h^t} \prod_{i=1}^k \frac{\partial h_{t-i+1}}{\partial h_{t-i}}$$ 
The backpropagated message is multiplied repeatedly by the gradient at each timestep in the scaffold. If the recurrence derivatives $\frac{\partial h_{i+1}}{\partial h_{i}}$ are large at some weight, the correspondingly larger backpropagated gradient $\frac{\partial e^t}{\partial h_{t-k}}$ will accelerate descent at that parameter. 
In other words, an unpredictable scaffold associated with a high error will dominate the gradient's sum over recurrences, delaying the acquisition of the symbol-matching rule. In the case of an LSTM, \citet{kanuparthi_h-detach:_2018} expressed the backpropagated gradient as an iterated addition of the error from each timestep, leading to a similar effect. 

See Appendix~\ref{sec:bptt} for confirmation of the difference in gradients between familiar and unfamiliar scaffolds. The speed of acquisition of the dependency rule in a familiar-scaffold training environment therefore has an explanation other than hierarchical composition. Therefore, in order to confirm our proposed compositional bias, we observe the interactions between scaffold and superstructure (long distance dependency) using DI.

\begin{figure}
    \centering
    \includegraphics[width=0.48\textwidth]{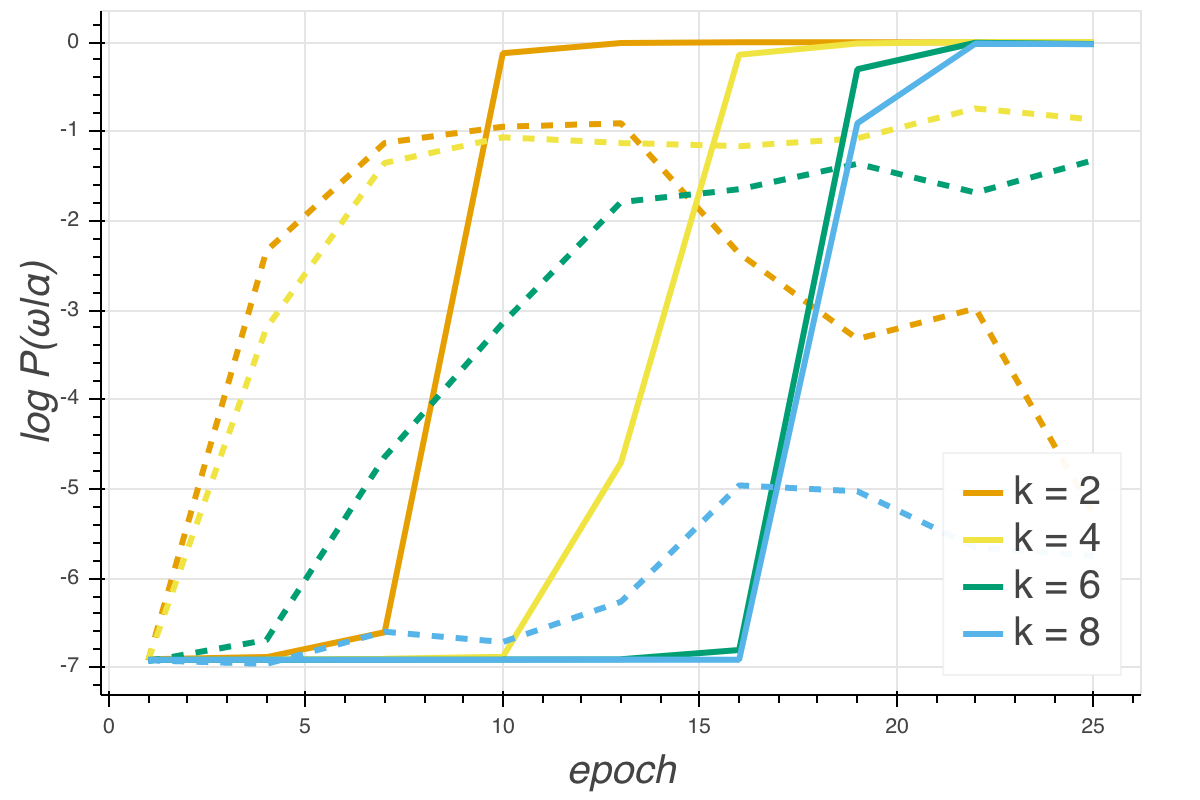}
    \caption{Mean target probability of $\omega$ at its correct timestep based on CD with $\alpha$ in focus, on out-domain test set. Solid lines are trained in the unfamiliar-scaffold set, dashed lines on familiar-scaffold.}
    \label{fig:cd_alpha}
\end{figure}

\subsubsection{Isolating the Effect of the Open-Symbol} \label{sec:isolating}
 
Raw predictions in the out-domain test setting appear to suggest that the familiar-scaffold training setting fails to teach the model to associate $\alpha$ and $\omega$. However, the changing domain makes this an unfair assertion: the poor performance may be attributed to wholesale memorization of $\alpha q$. To illustrate that the rule is learned regardless of training scaffolds, we use CD to isolate  the  contributions of the open symbol in the out-domain test setting (Figure~\ref{fig:cd_alpha}). Furthermore, we confirm that the familiar-scaffold training setting enables earlier acquisition of this rule.

\begin{figure}
    \centering
    \includegraphics[width=0.48\textwidth]{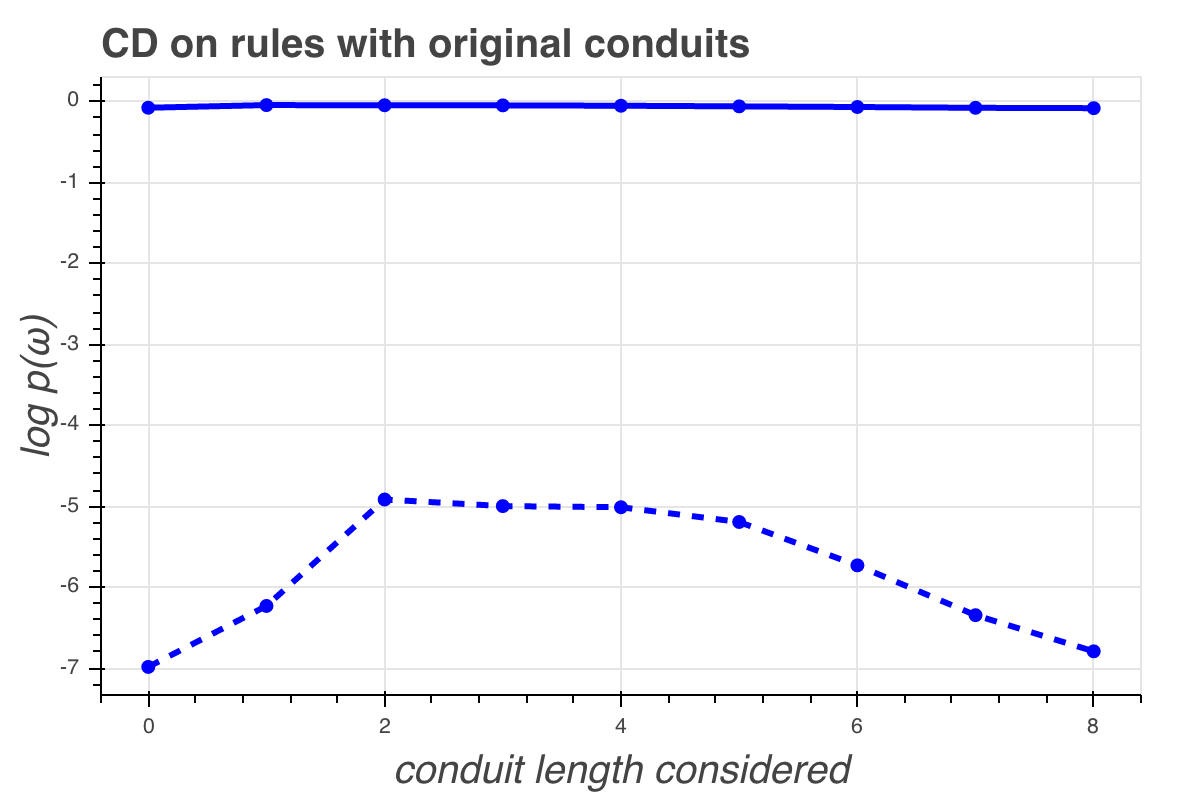}
    \caption{The predicted $P(x_{t} = \omega | x_{t-k} \ldots x_{t-k+i})$ according to CD, varying $i$ as the x-axis and with $x_{t-k} = \alpha$ and $k=8$. Solid lines are trained in the unfamiliar-scaffold set, dashed lines on familiar-scaffold.}
    \label{fig:incremental}
\end{figure}

To what, then, can we attribute the failure to generalize out-domain? Figure~\ref{fig:incremental} illustrates how the unfamiliar-scaffold model predicts the close symbol $\omega$ with high probability based only on the contributions of the open symbol $\alpha$. Meanwhile, the familiar-scaffold model probability increases substantially with each symbol consumed until the end of the scaffold, indicating that the model is relying on interactions between the open symbol and the scaffold rather than registering only the effect of the open symbol. Note that this effect cannot be because the scaffold is more predictive of $\omega$. Because each scaffold appears frequently outside of the specific context of the rule in the familiar-scaffold setting, the scaffold is \textit{less} predictive of $\omega$ based on distribution alone.

\begin{figure}
    \centering
    \includegraphics[width=0.48\textwidth]{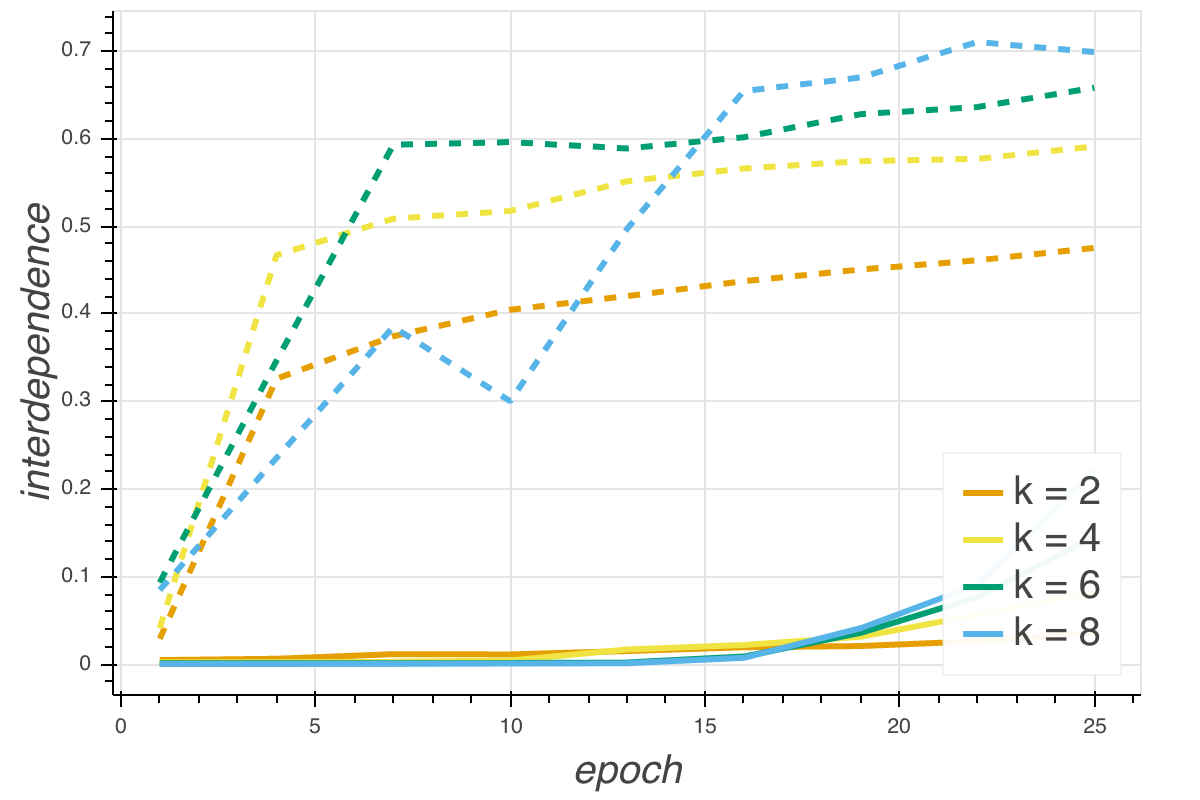}
    \caption{Mean $\textrm{DI}(\alpha, \textrm{scaffold})$ on the in-domain test set. Solid lines are trained in the unfamiliar-scaffold set, dashed lines on familiar-scaffold.}
    \label{fig:interdependence}
\end{figure}

These results indicate that predictable patterns play a vital role in shaping the representations of symbols around them by composing in a way that cannot be easily linearized as a sum of the component parts. In particular, as seen in Figure~\ref{fig:interdependence}, the DI between open symbol and scaffold is substantially higher for the familiar-setting model and increases throughout training. Long-range connections are not learned independently from scaffold representations, but are \textit{built compositionally} using already-familiar shorter subsequences as scaffolding.

{}

\section{Discussion \& Related Work}

Humans learn by memorizing short rote phrases and later mastering   the ability to construct deep syntactic trees from them \citep{lieven2008children}. LSTM models learn by backpropagation through time, which is unlikely to lead to the same \textit{inductive biases}, the assumptions that define how the model generalizes from its training data. It may not be expected for an LSTM to exhibit similarly compositional learning behavior by building longer constituents out of shorter ones during training, but we present evidence in favor of such learning dynamics.

LSTMs have the theoretical capacity to encode a wide range of context-sensitive languages, but in practice their ability to learn such rules from data is limited \citep{weiss-etal-2018-practical}. Empirically, LSTMs encode the most recent noun as the subject of a verb by default,  but they are still capable of learning to encode grammatical inflection from the first word in a sequence rather than the most recent \cite{DBLP:journals/corr/abs-1903-06400}. Therefore, while inductive biases inherent to the model play a critical role in the ability of an LSTM to learn \textit{effectively}, they are neither necessary nor sufficient in determining what the model \textit{can} learn. Hierarchical linguistic structure may be learned from data alone, or be a natural product of the training process, with neither hypothesis a foregone conclusion. We provide a more precise lens on how LSTM training is itself compositional, beyond the properties of data.


There is a limited literature on compositionality as an inductive bias of neural networks. 
\citet{saxe_mathematical_2019} explored how hierarchical ontologies are learned by following their tree structure in 2-layer feedforward networks. LSTMs also take advantage of some inherent trait of language~\citep{liu_lstms_2018} . The compositional training we have explored may be the mechanism behind this biased representational power.

Synthetic data, meanwhile, has formed the basis for analyzing the inductive biases of neural networks and their capacity to learn compositional rules. Common synthetic datasets include the Dyck languages~\citep{suzgun-etal-2019-lstm,skachkova-etal-2018-closing}, SPk~\citep{mahalunkar-kelleher-2019-multi}, synthetic variants of natural language~\citep{DBLP:journals/corr/abs-1903-06400,liu_lstms_2018}, and others~\citep{DBLP:journals/corr/abs-1906-00180,livska2018memorize,korrel-etal-2019-transcoding}. Unlike these works, our synthetic task is not designed primarily to test the biases of the neural network or to improve its performance in a restricted setting, but to investigate the internal behavior of an LSTM in response to memorization.

Investigations into learning dynamics like ours may offer insight into selecting training curricula.
The application of a curriculum is based on the often unspoken assumption that the representation of a complex pattern can be reached more easily from a simpler pattern. However, we find that effectively representing shorter scaffolds actually makes a language model \textit{less} effective at generalizing a long-range rule, as found by \citet{zhang_empirical_2018}. This less generalizable representation is still learned faster, which may be why \citet{zhang_boosting_2017} found higher performance after one epoch. Our work suggests that measures of length, including syntactic depth, may be inappropriate bases for curriculum learning.

\section{Future Work}

While we hope to isolate the role of long range dependencies through synthetic data, we must consider the possibility that the natural predictability of language data differs in relevant ways from the synthetic data, in which the scaffolds are predictable only through pure memorization. Because LSTM models take advantage of linguistic structure, we cannot be confident that predictable natural language exhibits the same cell state dynamics that make a memorized scaffold promote or inhibit long-range rule learning. Future work could test our findings on the learning process through carefully selected natural language, rather than synthetic, data. 

Our natural language results could lead to DI as a structural probe for testing syntax. Such a probe can be computed directly from an LSTM without learning additional parameters as required in other methods \citep{hewitt_structural_nodate}. In this way, it is similar to the probes that have been developed using attention distributions~\citep{clark_what_2019}. By computing associations naturally through DI, we can even escape the need to augment models with attention just to permit analysis, as \citet{smith_what_2017}.

Some effects on our natural language experiments may be due to the predictable nature of English syntax, which favors right-branching behavior. Future work could apply similar analysis to other languages with different grammatical word orders.

\section{Conclusions}


Using our proposed tool of Decompositional Interdependence, we illustrate how information exchanged between words aligns roughly with syntactic structure, indicating LSTMs compose meaning bottom-up. Synthetic experiments then illustrate that a memorized span intervening between a long distance dependency promotes early learning of the dependency rule, but fails to generalize to new domains, implying that these memorized spans are used as scaffolding in a bottom-up learning process. 

This combination of behaviors is similar to a syntactic language model, suggesting that the LSTM's demonstrated inductive bias towards hierarchical structures is implicitly aligned with our understanding of language and emerges from its natural learning process.

\section*{Acknowledgements}
We thank
Ida Szubert,
Annabelle Michael Carrell,
Seraphina Goldfarb-Terrant,
Craig Innes,
Kate McCurdy,
Yevgen Matusevych,
Andreas Grivas,
Nikolay Bogoychev,
Sameer Bansal,
Matthew Summers,
and Denis Emelin
for comments on early drafts of this paper.

\bibliography{acl2019}
\bibliographystyle{acl_natbib}

\clearpage
\newpage
\appendix

\section{Details of Contextual Decomposition} \label{sec:cd_details}

For an output hidden state vector $h^t$, CD will decompose it into two vectors: the relevant $h^t_{\beta}$, and irrelevant $h^t_{\bar{\beta}; \beta \interaction \bar{\beta}}$, such that:
$$h \approx h^t_{\beta} + h^t_{\bar{\beta}; \beta \interaction \bar{\beta}}$$

This decomposed form is achieved by linearizing the contribution of the words in focus at each gate. This is necessarily approximate, because the internal gating mechanisms in an LSTM each employ a nonlinear activation function, either $\sigma$ or tanh.  \citet{murdoch_beyond_2018} use a linearized approximation $L_{\sigma}$ for $\sigma$ and linearized approximation $L_{\tanh}$ for $\tanh$ such that for arbitrary input $\sum_{j=1}^{N} y_j$: 
\begin{equation}\label{eqn:linear}
\sigma{\left(\sum_{j=1}^{N} y_j\right)} = \sum_{j=1}^{N} L_{\sigma}(y_j)
\end{equation}

These approximations are then used to split each gate into components contributed by the previous hidden state $h^{t-1}$ and by the current input $x^t$, for example the input gate $i^t$:
\begin{eqnarray*} 
i^t &=& \sigma(W_i x^t + V_t h^{t-1} + b_i)\\
&\approx& L_{\sigma}(W_i x^t) + L_{\sigma}(V_t h^{t-1}) + L_{\sigma}(b_i)
\end{eqnarray*}{}

The linear form $L_{\sigma}$ is achieved by computing the Shapley value~\citep{shapley1953value} of its parameter, defined as the average difference resulting from excluding the parameter, over all possible permutations of the input summants. To apply Formula~\ref{eqn:linear} to $\sigma{(y_1 + y_2)}$ for a linear approximation of the isolated effect of the summant $y_1$:
$$
L_{\sigma}(y_1) = \frac{1}{2} [(\sigma(y_1) - \sigma(0)) + (\sigma(y_2 + y_1) - \sigma(y_1)) ]
$$

With this function, we can take a hidden state from the previous timestep, decomposed as $h^{t-1} \approx h^{t-1}_{\beta} + h^{t-1}_{\bar{\beta}; \beta \interaction \bar{\beta}}$ and add $x^t$ to the appropriate component. For example, if $x^t$ is in focus, we count it in the relevant function inputs when computing the input gate:
\begin{eqnarray*}
i^t &=& \sigma(W_i x^t + V_t h^{t-1} + b_i)\\
&\approx& \sigma(W_i x^t + V_t (h^{t-1}_{\beta} + h^{t-1}_{\bar{\beta}; \beta \interaction \bar{\beta}}) + b_i)\\
&\approx& [L_{\sigma}(W_i x^t + V_t h^{t-1}_{\beta}) + L_{\sigma}(b_i)]\\ 
&&+ L_{\sigma}(V_t h^{t-1}_{\bar{\beta}; \beta \interaction \bar{\beta}})\\ 
&=& i^t_{\beta} + i^{t}_{\bar{\beta}; \beta \interaction \bar{\beta}}
\end{eqnarray*}{}

This provides an expression of the approximate input gate as the sum of relevant and irrelevant components. By ignoring the irrelevant components while computing the module output $h^t$, we produce $h^t_{\beta}$. Thus we linearize and isolate the effect of $\beta$.

\section{The Effect of Rule Frequency and Length} \label{sec:frequency}
\begin{figure*}
    \centering
    \includegraphics[width=0.32\textwidth]{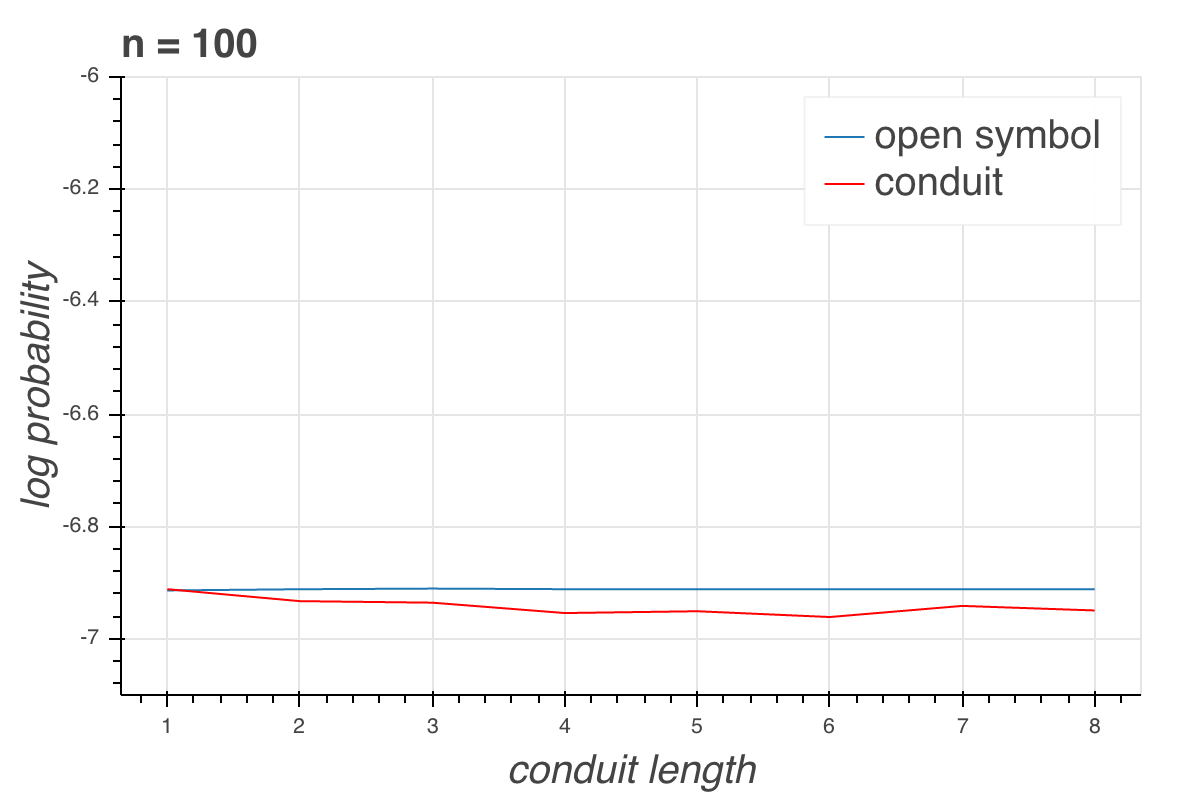}
    \includegraphics[width=0.32\textwidth]{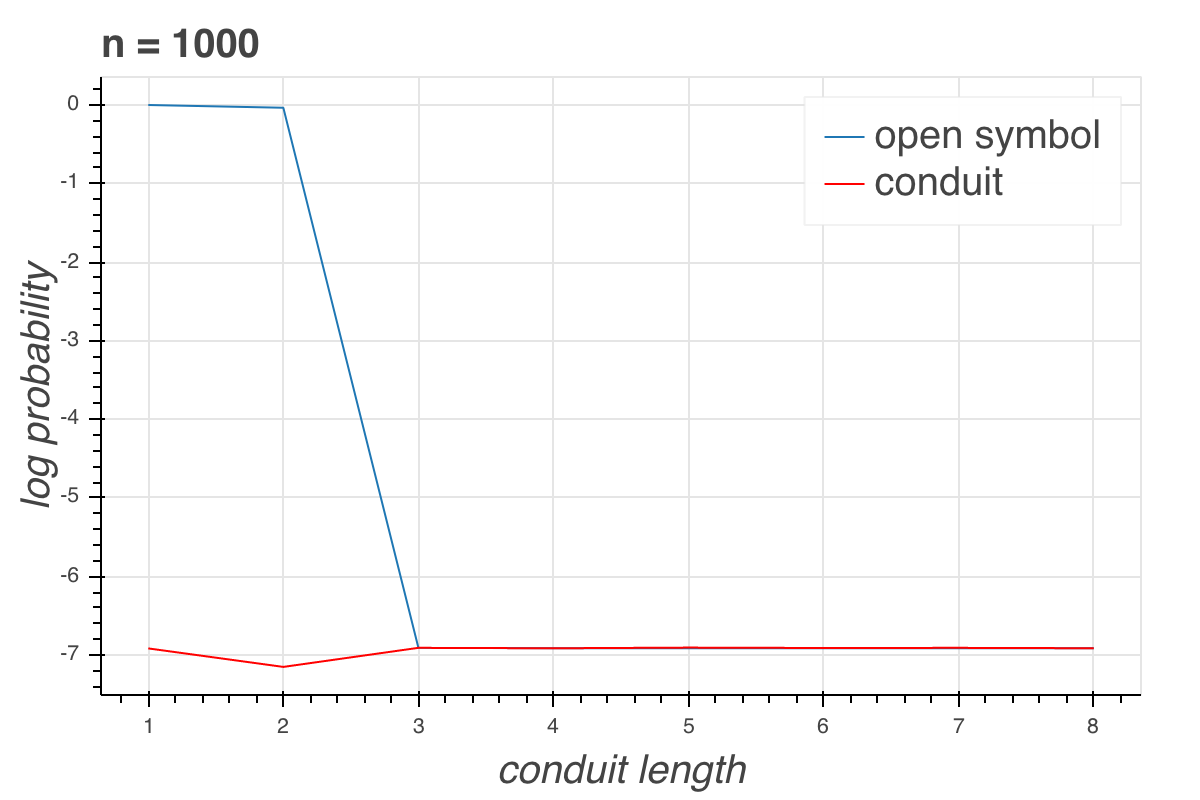}
    \includegraphics[width=0.32\textwidth]{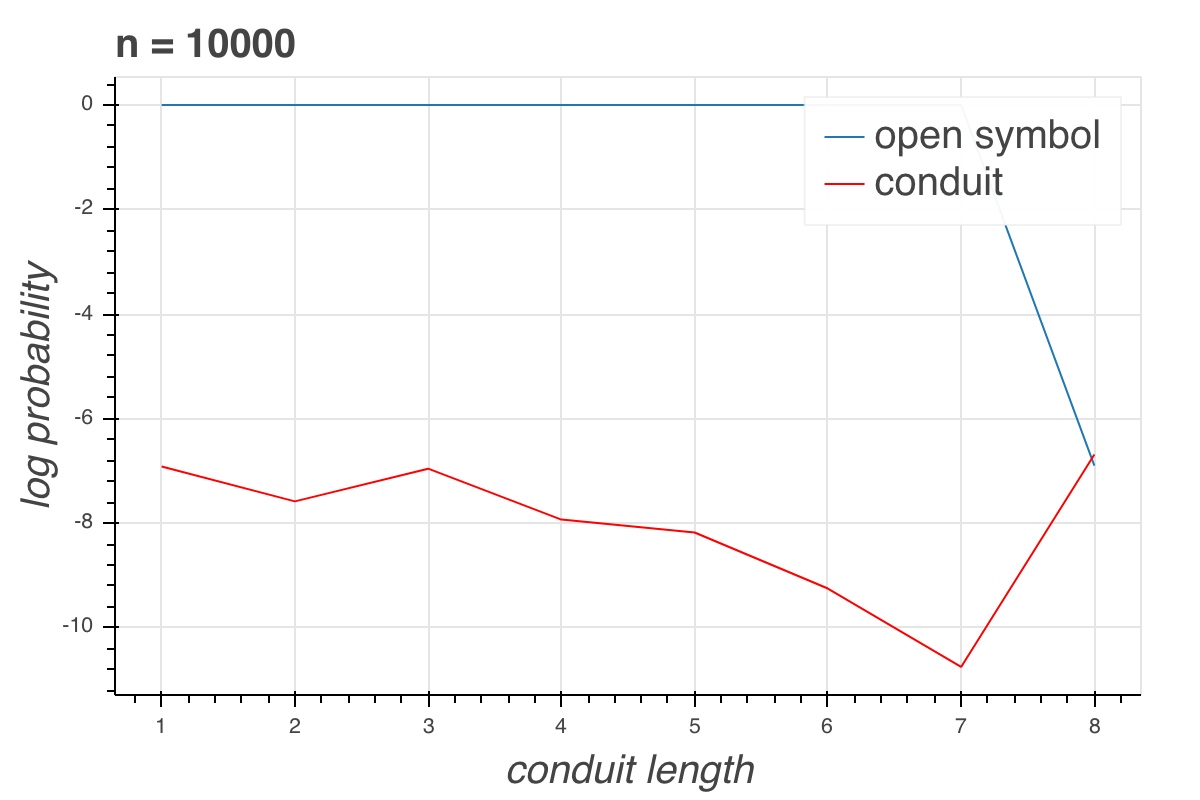}
    \caption{The predicted probability $P(x_{t} = \omega)$, according to the contributions of open symbol $x_{t-k} = \alpha$ and of the scaffold sequence $x_{t-k+1} \ldots x_{t-1}$, for various rule occurrence counts $n$. Shown at 40 epochs.}
    \label{fig:rule_frequency}
\end{figure*}

Here, we investigate how the frequency of a rule affects the ability of the model to  learn the rule by varying the number of rule occurrences $n$ and the rule length $k$. 

The results in Figure~\ref{fig:rule_frequency} illustrate how a longer scaffold length requires more  examples before the model can learn the corresponding  rule.  We consider the probability assigned to the close symbol according to the contributions of the open symbol, excluding interaction from any other token in the sequence. For contrast, we also show the extremely low probability assigned to the close symbol according to the contributions of the scaffold taken as an entire phrase. In particular, note the pattern when the rule is extremely rare:  The probability of the close symbol $\beta$ as determined by the open symbol $\alpha$ is low but steady, while the probability  as determined by the scaffold declines with scaffold length  due to the accumulated low probabilities from each element in the sequence.

\section{Smaller scaffold gradient, faster rule learning} \label{sec:bptt}

\begin{figure}
\includegraphics[width=\linewidth]{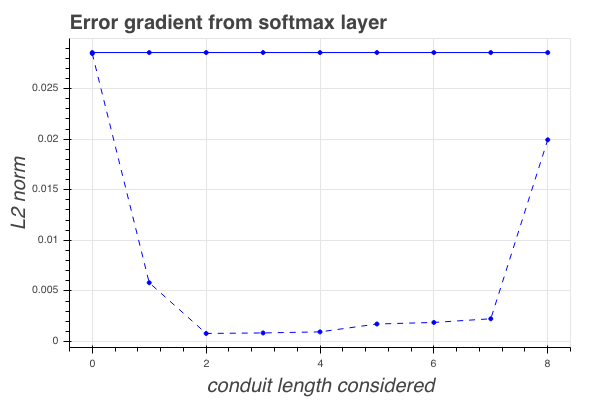}
\caption{Average gradient magnitude $\Delta E_{t+-k+d}$, varying $d$ up to the length of the scaffold. Solid lines are the unpredictable scaffold setting, dashed lines are the predictable scaffold setting.}
\label{fig:error}
\end{figure}

Figure~\ref{fig:error} confirms that a predictable scaffold is associated with a smaller error gradient. Because of the mechanics of backpropagation through time next described, this setting will teach the $\alpha/\omega$  rule faster.

\end{document}